\documentclass[runningheads]{llncs}

\usepackage{eccv}

\usepackage{eccvabbrv}

\usepackage{graphicx}
\usepackage{booktabs}
\usepackage{amssymb}
\usepackage{amsmath}
\usepackage{bm}

\usepackage[accsupp]{axessibility}  

\usepackage{indentfirst} 
\usepackage{multirow}
\usepackage{wrapfig}

\usepackage{hyperref}

\usepackage{orcidlink}
\newcommand*\samethanks[1][\value{footnote}]{\footnotemark[#1]}

\begin{document}

\title{Take A Step Back: Rethinking the Two Stages in Visual Reasoning} 

\titlerunning{Take A Step Back}

\author{Mingyu Zhang\inst{1,}\thanks{Equal contribution.}\orcidlink{0009-0009-4174-9601} \and
Jiting Cai\inst{1,}\samethanks\orcidlink{0009-0006-4716-8144} \and
Mingyu Liu\inst{2}\orcidlink{0009-0006-1379-2846} \and
Yue Xu\inst{1}\orcidlink{0000-0001-7489-7269} \and
Cewu Lu\inst{1}\orcidlink{0000-0003-1533-8576} \and
Yong-Lu Li\inst{1}\orcidlink{0000-0003-0478-0692}\thanks{Corresponding author.}
} 

\authorrunning{Zhang et al.}

\institute{Shanghai Jiao Tong University \\
\email{\{sjtuzmy2003, Caijiting, silicxuyue, lucewu, yonglu\_li\}@sjtu.edu.cn} \and
Zhejiang University \\
\email{mingyuliu@zju.edu.cn}}

\maketitle

\begin{abstract}
As a prominent research area, visual reasoning plays a crucial role in AI by facilitating concept formation and interaction with the world.
However, current works are usually carried out separately on small datasets thus lacking generalization ability.
Through rigorous evaluation of diverse benchmarks, we demonstrate the shortcomings of existing ad-hoc methods in achieving cross-domain reasoning and their tendency to data bias fitting.
In this paper, we revisit visual reasoning with a two-stage perspective: (1) symbolization and (2) logical reasoning given symbols or their representations. 
We find that the reasoning stage is better at generalization than symbolization.
Thus, it is more efficient to implement symbolization via \textbf{separated} encoders for different data domains while using a \textbf{shared} reasoner.
Given our findings, we establish design principles for visual reasoning frameworks following the separated symbolization and shared reasoning.
The proposed two-stage framework achieves impressive generalization ability on various visual reasoning tasks, including puzzles, physical prediction, and visual question answering (VQA), encompassing 2D and 3D modalities.
We believe our insights will pave the way for generalizable visual reasoning.
\textbf{Our code is publicly available at \url{https://mybearyzhang.github.io/projects/TwoStageReason}.}
\end{abstract}
\section{Introduction}
\label{sec:intro}

Reasoning ability~\cite{pearl2018book} is a concentrated embodiment of human intelligence, serving as the foundation for concept formation, cognitive understanding of the world, and interaction with the environment. 
Specifically, visual reasoning, being one of the primary modes through which humans acquire information and understanding, has been the focus of extensive research. 
In recent years, with the advancements in deep learning, numerous works~\cite{hong2021ptr, wen2020multi, Goyal_2017_CVPR, mcduff2022causalcity, yi2019clevrer, Mao2019NeuroSymbolic} on visual reasoning have been proposed. Additionally, various datasets~\cite{zhang2019raven, baradel2019cophy, janny2022FilteredCoPhy, Antol_2015_ICCV, Goyal_2017_CVPR, Johnson_2017_CVPR, Hudson_2019_CVPR} have also emerged to evaluate the reasoning models.

However, a notable limitation of existing visual reasoning works lies in their direct reliance on entangling the recognition and reasoning phrases via end-to-end deep learning models, \eg, recognizing the concepts in images while answering logical questions~\cite{zhang2019raven}. 
However, this paradigm has obvious limitations as follows:
1) Reasoning annotation (rule, relation) is much more costly and difficult than symbol annotation (triangle, cube, apple), thus current rigorous visual reasoning datasets are usually small. Therefore, current approaches tend to be heavily task-specific on small datasets~\cite{zhang2019raven, li2011comparing, zerroug2022benchmark}, hindering their generalization potential.
2) Pursuing a versatile model for symbol recognition and logical reasoning simultaneously may be inefficient and challenging. Even recent large language models (LLM) struggle with diverse visual reasoning tasks~\cite{zhao2023mmicl}.

\begin{figure}[t]
    \centering
    \includegraphics[width=0.99\textwidth]{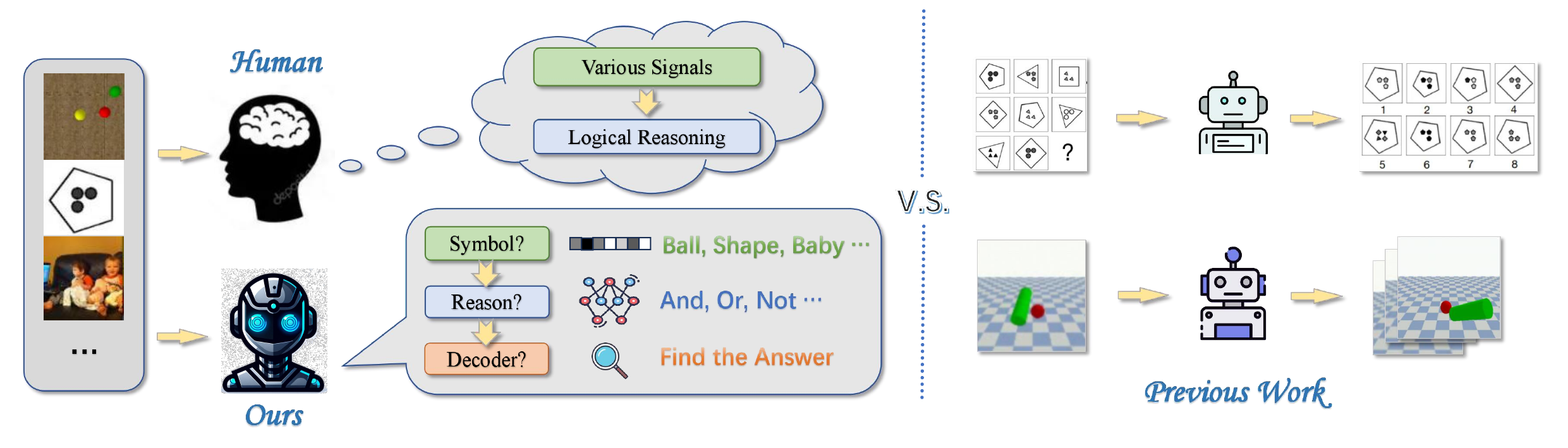}
    \caption{Comparison between end-to-end model, human, and our framework. Previous works usually use a specific end-to-end model for each task, while our framework shares a logical reasoner similar to human intelligence.}
    \label{fig:teaser}
\end{figure}

In this paper, we argue that visual reasoning is rooted in first grounding symbols derived from visual signals, followed by logical reasoning, as shown in Fig.~\ref{fig:teaser}.
Thus, a question arises: \textbf{should the two stages be entangled or disentangled?}
Reasoning naturally has a generalization property than symbolization.
For example, we use \textit{similar} logic to analyze the rules of different tasks (\eg, play go, do the math, and discover anomaly) but very \textit{different} knowledge to recognize the alphabet and objects.
Thus, we assume that disentangled symbolization and reasoning would be a more wise choice. 
The recent success of Large Language Models (LLM) on text-based reasoning tasks~\cite{tsai2023can} also validates this point, as LLMs directly leverage the abstract symbols (language) derived from human observation and focus on high-level linguistic tasks. 
Relatively, Multi-Modal Large Language Models (MLLM) still struggle with visual reasoning~\cite{gong2023multimodal} even with more parameters.
Recently, another related research trend is the neuro-symbolic method. 
The neuro-symbolic approach transforms raw inputs into explicit symbols for subsequent reasoning and analysis~\cite{yu2023survey}. 
However, neuro-symbolic methods often remain confined to a single dataset~\cite{amizadeh2020neuro, gupta2023visual, suris2023vipergpt}, making it challenging to achieve generalization across different tasks.

We conduct comprehensive experiments on various tasks of multiple benchmarks with significant domain gaps to verify our assumption.
We formulate the symbolization stage as the representation extraction with Deep Neural Networks (DNN) and implement the logical reasoner with various architectures (MLP, CNN, GNN, Transformer, Neuro-Symbolic models, LLM, \etc). 
We mainly investigate \textbf{two key questions}:
(1) Within a trained DNN, where does the symbolization phrase conclude? That is, identifying the suitable symbol (representations) for reasoning, such as the depth of the model, feature characteristics, \etc.
(2) Given the abstracted symbols, what types of models and training strategies are most suitable for reasoning and endowing the generalization capability?

For the \textbf{first} problem, we find that different tasks and domains need very different scales of parameters or model depths to achieve a good symbolization.  
Thus, for a specific domain, a small \textbf{separated} in-domain encoder is enough to extract symbols from the data for the subsequent reasoning phrase.
Though a general and large foundation model like CLIP~\cite{radford2021learning} can do well on some tasks, it still struggles on tasks with a huge domain gap with its training data~\cite{radford2021learning, liu2023improved}.
For the \textbf{second} problem, our results reveal that existing methods struggle to execute cross-domain reasoning, instead fitting biases aligned with training data. 
Thus, maybe we should or can only achieve a generalizable \textbf{shared} reasoner via training it on various reasoning tasks (puzzles, physical prediction, VQA) and data domains (2D, 3D, text), \ie, ``\textbf{approximation principle}''.

Building upon our findings, we build a concise framework with separated encoders to achieve the optimal symbolization for different data domains and a shared reasoner following the ``approximation principle''. 
Our method performs exceptionally well with fewer parameters across cross-domain benchmarks.

In general, our contributions are: 
(1) We conclude an efficient two-stage perspective for visual reasoning drawing inspiration from previous visual reasoning networks.
(2) We investigate the optimal design principles of symbolization and logical reasoning for visual reasoning.
(3) Accordingly, we introduce a concise framework with decent performance on multiple datasets with domain gaps.
\section{Related Work}

{\bf Visual Reasoning.}
Visual reasoning is a subfield of computer vision and artificial intelligence that aims to enable machines to reason about visual information in a human-like manner. 
It involves using machine learning techniques to analyze images or videos and making decisions based on that analysis. 
Several recent studies have explored various architectures for deep neural networks designed specifically for visual reasoning tasks such as the Visual Question Answering (VQA)~\cite{Hudson_2019_CVPR,Goyal_2017_CVPR}, 
video causal inference~\cite{baradel2019cophy,ocl}, dynamic prediction for 3D scenes~\cite{janny2022FilteredCoPhy}, \etc.
However, most of them are domain-specific and lack generalization ability.
These architectures have typically combined convolutional neural networks for visual feature extraction with recurrent neural networks or LSTM for language processing~\cite{yi2019clevrer}.
Another area of research in visual reasoning is the development of knowledge-based systems~\cite{wen2020multi}, which rely on structured representations of knowledge about the world to reason about visual information. 
For example, Hong~\etal~\cite{hong2021ptr} built a system that could reason about the layout of objects in a scene or the relationships between different objects in an image.
However, the essence of reasoning itself and the certain stages of visual reasoning, which is a basic problem in this domain, remains unclear.

{\bf Neuro-Symbolic Learning.}
Neuro-symbolic learning~\cite{garcez2022neural, yu2023survey,zhang2021neural,garcez2015neural,hakev2,yi2018neural,gupta2023visual,Symbol-LLM,suris2023vipergpt} combines symbolic reasoning with neural networks to address complex problems in domains like computer vision~\cite{hakev2,kroshchanka2021neural,yu2022probabilistic}, natural language processing~\cite{hamilton2022neuro,liu2022neural}, and knowledge inference~\cite{lemos2020neural,cornelio2022learning}. Neuro-symbolic methods are categorized into three types~\cite{yu2023survey}: 1) \textit{Learning for reasoning} uses neural systems to enhance symbolic reasoning; 2) \textit{Reasoning for learning} uses symbolic systems to aid neural learning and improve interpretability; and 3) \textit{Learning-reasoning} involves bidirectional interaction between neural and symbolic systems. Wu~\etal~\cite{Mao2019NeuroSymbolic} and Gupta~\etal~\cite{gupta2023visual} have proposed notable neuro-symbolic approaches for image feature extraction and problem-solving. Despite its promise, neuro-symbolic integration has limitations, particularly in generalizing reasoning abilities across tasks~\cite{yu2023survey}. Our framework differs from neuro-symbolic methods in two key ways: (1) It includes both \textit{end-to-end} and \textit{neuro-symbolic models}, with the latter as a subset; (2) It can explain the generalization of the reasoner, which neuro-symbolic methods cannot.

{\bf Visual Reasoning Benchmarks.} 
Researchers have developed various cross-modal visual reasoning benchmarks to evaluate reasoning models, including 2D puzzles, 3D physical prediction, Visual Question Answering (VQA), \etc.
2D puzzle datasets explore relationships among visual elements. Notable datasets include RAVEN~\cite{zhang2019raven}, SVRT~\cite{li2011comparing}, CVR~\cite{zerroug2022benchmark}, Bongard-HOI~\cite{jiang2022bongard}, and Bongard-LOGO~\cite{nie2020bongard}.
Intuitive physics datasets, such as CoPhy~\cite{baradel2019cophy}, Filtered-CoPhy~\cite{janny2022FilteredCoPhy}, Space~\cite{duan2021space}, and Space++\cite{duan2022pip}, focus on how humans perceive and reason about the physical world.
VQA requires machines to understand both natural language questions and images~\cite{WU201721}, with datasets including VQAv1~\cite{Antol_2015_ICCV}, VQAv2~\cite{Goyal_2017_CVPR}, CLEVR~\cite{Johnson_2017_CVPR}, and GQA~\cite{Hudson_2019_CVPR}.
\section{Preliminary}

In this section, we first formulate visual reasoning following the proposed two-stage view.

\subsection{Two Stages}
As described above, visual reasoning can be divided into two stages: the symbolization stage extracts symbolic representations of the underlying data, and the reasoning stage performs logical reasoning.

For humans, different modalities of visual and auditory information collected from our sensors are converted into electrical signals through different pathways and then sent to the cerebellar cortex to perform logical reasoning~\cite{funamizu2016neural}.
Analogously, separated task-specific symbolizers and a shared domain-independent reasoner would be a reasonable choice for a general visual reasoning machine.
Besides, the reasoner should be capable of performing unified reasoning on input information from various modalities. 
In other words, \textit{the essence of reasoning lies in its generalization ability}.

{\bf Symbolization Stage.}
During the stage of symbolization, we implement various task-oriented feature extraction networks. These networks employ symbol encoders tailored for each task, transforming multi-modal inputs (text, image, video) into symbol representations. 
Formally, suppose we have $n$ tasks. 
For the $i$-th task, we have the input data $\bm{x}^{i}$ and the task $t^{i}$, and the task-oriented encoder $E^{i}$. Then we get the symbol representation set $\bm{f}^{i}$ via:

\begin{equation}
    \bm{f}^{i} = E_i(\bm{x}^{i} \mid t^{i}).
\end{equation}

{\bf Reasoning Stage.}
The reasoner is fed by symbolic representations for each specific task, in a bid to capture a deeper and more comprehensive understanding of the underlying patterns and relationships embedded within the data. 
For symbol representation sets $\{\bm{f}^{(i)}\}^n_{i=1}$ of all tasks, we send them into the reasoner $R$, and get its reasoning result set $\{\bm{c}^{i}\}^n_{i=1}$ after the logic processing to facilitate problem-solving across various modalities:
\begin{equation}
    \{\bm{c}^{i}\}^n_{i=1} = R(\{\bm{f}^{i}\}^n_{i=1}).
    \label{eq:reasoning}
\end{equation}

{\bf Task-specific Heads.}
The final part of our framework is the task-specific heads, which take the reasoning results from the reasoner as input and generate task-specific answers. 
For different tasks, we need to construct task-specific classification or regression heads $H_i$ to get the final output $s^i$. That says:
\begin{equation}
    s^i = H_i(\bm{c}^{i} \mid t^{i}).
\end{equation}
Next, we can compare the output with the ground truth to compute gradients for training the entire framework.

\section{Symbolization-Reasoning Framework}
\subsection{Entanglement v.s. Disentanglement}

Given the two stages, a natural question arises: should the symbol encoder (symbolization) and reasoner (reasoning) be shared or separated for tasks?

\begin{wrapfigure}{r}{0.55\textwidth}
  \begin{center}
    \includegraphics[width=0.55\textwidth]{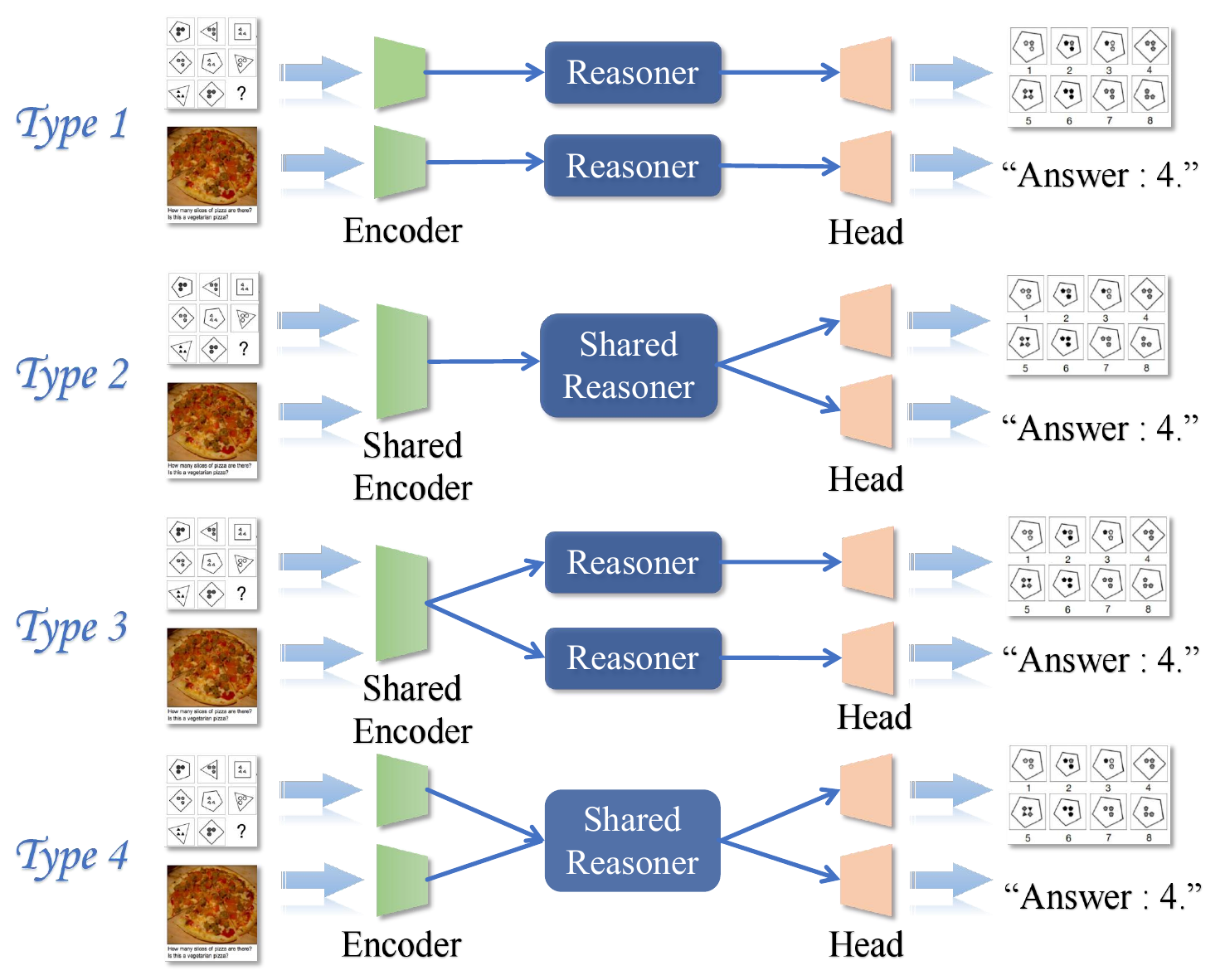}
  \end{center}
  \caption{Entanglement v.s. Disentanglement. Type 1: the symbol encoder and reasoner are all separated; Type 2: both the encoder and reasoner are shared; Type 3: only the encoder is shared; Type 4: only the reasoner is shared.}
    \label{fig:enter-label1}
\end{wrapfigure}

To validate our \textit{shared reasoner only} assumption, we conduct a comparison between different designs (Fig.~\ref{fig:enter-label1}):
1) \textbf{Both-Separated}: the symbol encoder and logical reasoner are all separated (ad-hoc models for each task);
2) \textbf{Both-Shared}: both the encoder and reasoner are shared.
3) \textbf{Shared-Encoder-Only}: only the symbol encoder is shared;
4) \textbf{Shared-Reasoner-Only}: only the reasoner is shared.

We compare the above four designs on several multi-modal visual reasoning benchmarks~\cite{zhang2019raven, zerroug2022benchmark, li2011comparing, jiang2022bongard, yang2023neural}. 
For the shared encoder/reasoner, we adopt more 
parameters 
to balance them with the sum of parameters of the separated encoders/reasoners.
In the experiments of Sec.~\ref{sec:exp}, we find that the Shared-Reasoner-Only (Type 4) outperforms Shared-Encoder-Only (Type 3) and Both-Shared (Type 1 and 2) a lot on all benchmarks. Besides, the Shared-Reasoner-Only even beats the ad-hoc Both-Separated on some benchmarks, validating its superiority and generalization ability.

\subsection{Symbolization Depth}

Next, we probe the appropriate depth of the symbol encoder for different tasks.
The symbolization stage involves processing inputs from different domains and mapping them to the conceptual level, \ie, symbols.
Though we can use binary or index-like (one-hot) representations for symbols, in the context of deep learning, we choose the more representative way, \ie, high-dimensional features extracted from DNN.
Intuitively, different tasks may need different levels of symbolization,
Then, the next question is how to determine the level of abstraction for each task.

To answer this question, we design experiments to quantitatively probe the degree of symbolization in the process. 
To control variables, we employ the same feature extraction network (ResNet~\cite{he2016deep}) for cross-domain tasks while continuously adjusting the network depth. At different depths, we connect the output
of the symbolization network to the same reasoner and measure the \textit{accuracy} as an indicator of symbolization completion. 

\begin{wrapfigure}{r}{0.55\textwidth}
    \begin{center}
      \includegraphics[width=0.55\textwidth]{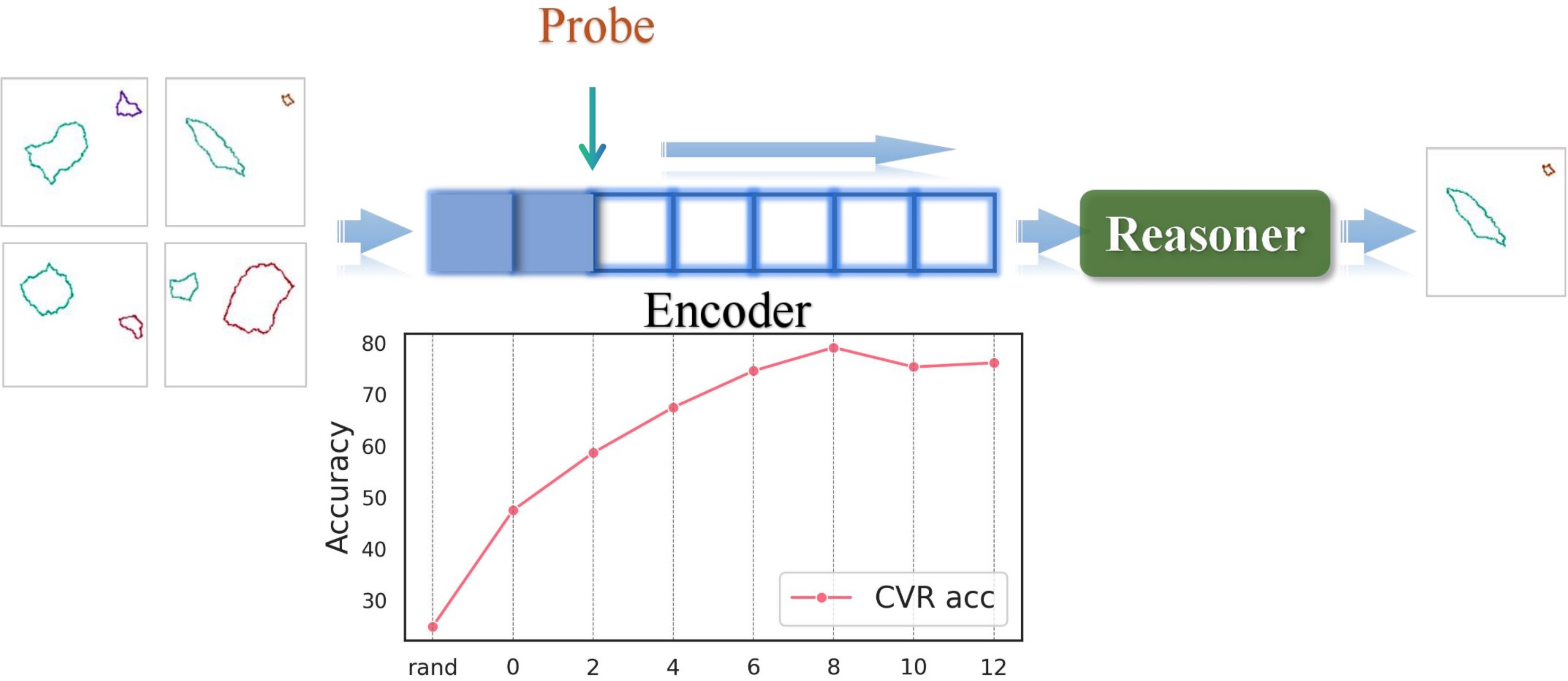}
    \end{center}
    \caption{Probing process of symbolization. We vary the depths of the symbol encoder (ResNet) and train the framework while recording the accuracy at each encoder depth. An inflection point occurs in the curve at moderate depths.}
    \label{fig:enter-label3}
\end{wrapfigure}

We hypothesize that when symbolization is complete, the network depth-accuracy curve will exhibit a distinct inflection point. By monitoring the occurrence of this inflection point, we can select the appropriate depth for the reasoner for different tasks, as shown in Fig.~\ref{fig:enter-label3}.
In experiments, we find the result consistent with our common sense:
both excessively shallow and deep networks are detrimental to reasoning tasks. 
On one hand, if the encoder is too shallow, symbolization will not be fully achieved before being inputted into the reasoner,
while the reasoner would have to conduct part of the symbolization work, thus impacting the performance of reasoning. 
On the other hand, deeper networks tend to overfit to one single task, thus weakening the performance of the shared reasoner which aims for generalization.
The symbolization depths also vary for tasks in different domains. Improperly setting the depth leads to conflicts in the shared parameters, resulting in the poor performance of deeper layers.

\subsection{Reasoner Architecture}
Next, we want to figure out which architecture is more suitable for the reasoner, which is a problem of a long history. Many works have been proposed and achieved improvements on various visual reasoning benchmarks~\cite{zhang2019raven, jiang2020defense, Goyal_2017_CVPR}.
Here, each task is handled by its respective encoder and task head, designed to adapt to the inherent characteristics of its data. We use a \textit{shared reasoner} for all tasks and take into the symbol representations following Eq.~\ref{eq:reasoning}.

We choose a line of architectures as the reasoner candidates:
Multilayer Perceptron (MLP), Convolutional Neural Networks (CNN), and Transformer that achieve great success across numerous tasks.
We also explore a hybrid neuro-symbolic model~\cite{Mao2019NeuroSymbolic} combining the representational capacity of neural networks with the interpretability of symbolic systems.
Furthermore, we adopt the popular graphical and autoregressive models:
Graph Convolutional Networks (GCN)~\cite{gori2005new} and MiniGPT-4~\cite{zhu2023minigpt,chen2023minigptv2}.
The above models afford us a comprehensive and diverse range of methodologies.
If a robust and consistent performance is witnessed across various domain-diverse datasets, it leads us to hypothesize the presence of a certain type of architecture that excels specifically at logical reasoning.

\subsection{Generalization of Reasoner}
\begin{wrapfigure}{r}{0.55\textwidth}
    \centering
    \includegraphics[width=0.55\textwidth]{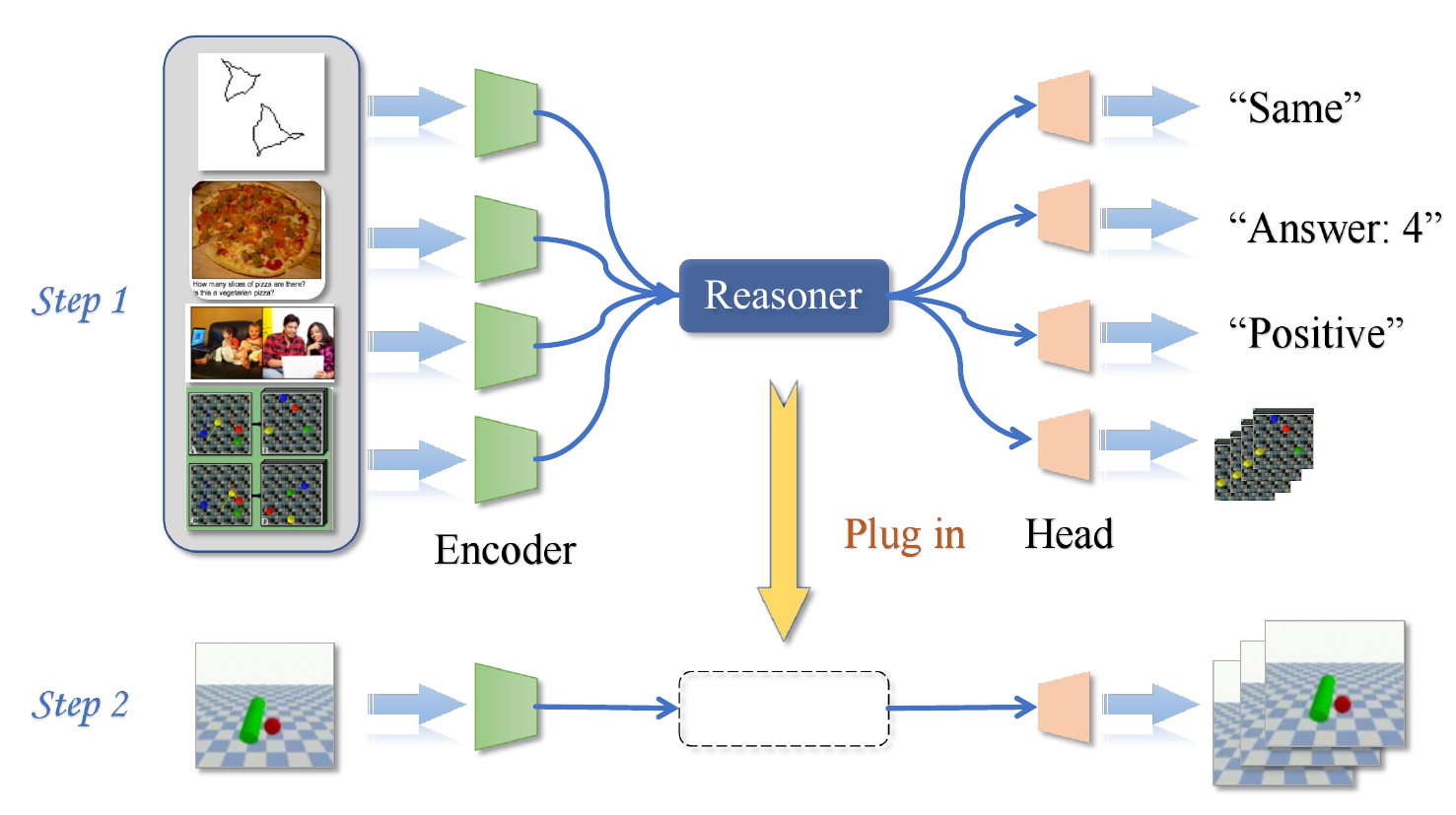}
    \caption{``Approximation principle'' verification with a shared reasoner. In step 1, the process entails the selection of 1-4 datasets, namely SVRT, Bongard-HOI, CoPhy-Balls, and VQAv2, to train the reasoner. This combination offers a total of 15 possible permutations. In step 2, the proficiently trained reasoner is subjected to rigorous testing on the CoPhy-Collision dataset for evaluation and validation purposes.}
    \label{fig:enter-label4}
\end{wrapfigure}

Last but not least, we aim to verify our ``approximation principle'', \ie, a good reasoner affording generalizable logical reasoning ability can be approached by training it with diverse tasks and data from diverse domains.
We believe that reasoning connotes universality and generalization. 
Therefore, 
we initially train a complete two-stage model on one task, and then directly take its reasoner and pair it with another symbol encoder of another task. If the reasoner has generalization ability, it should suit the encoder of the other tasks well.
However, in our test, a reasoner trained with only one task/domain usually generalizes not well.
Thus, we next verify whether the generalization ability of the reasoner is better given the training on more tasks/domains, as shown in Fig.~\ref{fig:enter-label4}.

We find that, as more and more data from different tasks and domains are involved, the overall task becomes increasingly challenging.
However, the reasoner will concentrate on ``pure'' reasoning instead of task/domain-specific solving thus endowing better generalization ability (detailed in Sec.~\ref{sec:exp}).
That says, our ``approximation principle'' is reasonable.
Thus, we can predict that the reasoner should perform better on out-of-domain tasks as the training data and task increase.
Moreover, a shared and cross-task trained reasoner makes the whole visual reasoning framework lighter and more efficient.
\section{Experiments}
\label{sec:exp}
In this section, we present our experimental settings and empirical results to support our assumptions. Our experiments are designed to answer the following questions:

\noindent{\bf (1)} Are there any differences in transferability between the symbolization and reasoning stages?

\noindent{\bf (2)} Given the presence of two stages, what is the impact of different levels of symbolization depth, and can we identify a clear indicator for the termination of symbolization?

\noindent{\bf (3)} Which architectures suit the reasoning best and can the ``approximation principle'' guide the training of the reasoner to achieve better generalization?

\subsection{Dataset and Setting}
{\bf 2D Puzzles.}
We use RAVEN~\cite{zhang2019raven}, CVR~\cite{zerroug2022benchmark}, SVRT~\cite{yang2023neural}, Bongard-LOGO~\cite{nie2020bongard}, and Bongard-HOI~\cite{jiang2022bongard} as our 2D datasets to assess the reasoning ability of neural networks.
CVR and SVRT focus on the attributes like shape, color, size, \etc, of geometric objects.
RAVEN, beyond that, needs to deal with positional and logical relationships such as AND, OR, XOR, \etc.
Bongard-LOGO and Bongard-HOI follow the rules of the Bongard problem, aiming to find the deep common features and concepts among contrastive samples.
More details of these datasets will be elaborated in the Supplementary.

{\bf 2D VQA.}
We choose VQAv2~\cite{Goyal_2017_CVPR}, which is a bias-free and balanced dataset. 
Built upon COCO, VQAv2~\cite{Goyal_2017_CVPR} carefully selects questions and images to ensure an equal distribution of answers for each question type. 
It consists of 1,000 natural language descriptions, each question having ten options from which the model needs to make a selection.

{\bf 3D Intuitive Physics.}
We select Filtered-CoPhy~\cite{janny2022FilteredCoPhy}. 
It involves three tasks: 
block tower, reasoning about the stability of stacked blocks; 
balls, reasoning about the motion rules of spherical objects; 
and collision, reasoning about the physical phenomena during object collisions. 
As we focus on the reasoning more instead of frame prediction, here we only utilize the differences between scene keypoints as the measurement.

{\bf Implementation Details.}
All the networks are trained for within 100 epochs using the Adam optimizer~\cite{kingma2014adam}.
The learning rate and weight decay are finetuned with the assistance of Optuna~\cite{akiba2019optuna}.
For our \textbf{lite} reasoner, we choose modules from MLP, CNN, Transformer, \etc, with parameters \textit{no more than 100M}.
To test the generalization of the reasoner, we freeze the reasoner network and train the encoder and head modules.
All experiments were conducted on 4 Titan XP GPUs except for the experiment with LLM performed on 1 V100-32GB GPU.
More details of network structure and hyperparameter configuration of each setting and model will be elaborated in the Supplementary.

\subsection{Entanglement v.s. Disentanglement Analysis}
To compare the three types in Fig.~\ref{fig:enter-label1}, we 
train the models using five datasets: RAVEN~\cite{zhang2019raven}, CVR~\cite{zerroug2022benchmark}, SVRT~\cite{li2011comparing}, Bongard-HOI~\cite{jiang2022bongard}, and Bongard-LOGO~\cite{nie2020bongard}.
To control variables and facilitate model sharing, for all types above we adopt ResNet-18~\cite{he2016deep} as the encoder and an MLP as the reasoner.

\begin{table*}[t]
  \centering
  \caption{Entanglement v.s. Disentanglement. We demonstrate the impact of using shared modules in the two-stage framework, \ie, both separated encoder and reasoner, both shared, shared encoder and shared reasoner as shown in Fig.~\ref{fig:enter-label1}.
  }
  \resizebox{0.8\textwidth}{!}{
  \begin{tabular}{c|ccccc}
     & \textbf{RAVEN} & \textbf{CVR} & \textbf{SVRT} & \textbf{Bongard-HOI} & \textbf{Bongard-LOGO} \\
    \midrule
    Both-Separated (ad-hoc) & 53.40 & 74.04 & 86.20 & 61.64 & 72.16 \\ \hline
    Both-Shared & 35.86 & 55.48 & 51.58 & 52.65 & 60.83  \\ 
    Shared-Encoder-Only & 42.06 & 54.82 & 58.24 & 52.65 & 63.17 \\
    Shared-Reasoner-Only & \textbf{55.72} & \textbf{72.40} & \textbf{86.64} & \textbf{57.65} & \textbf{69.64} \\
  \end{tabular}}  
  \label{tab:three_design}
\end{table*}

The results are depicted in Tab.~\ref{tab:three_design}, we find that the performance of the Shared-Reasoner-Only is comparable with the ad-hoc Both-Separated on all five datasets and even higher on RAVEN and SVRT. Besides, Shared-Encoder-Only and Both-Shared show obvious inferior performance on all datasets.
This reflects the validity of our design in employing task-specific symbolizers and shared reasoners across multiple tasks.

\begin{table*}[t]
    \centering
    \caption{Detailed results of the impact of varying encoder depths in the symbolization stage.  Within a certain range, different tasks exhibit an improvement in scores as the symbol encoder deepens, however, each task displays a different inflection point.}
    \resizebox{1.0\textwidth}{!}{
        \begin{tabular}{c|cccccccc}
            & Random & ResBlock 0 & ResBlock 2 & ResBlock 4 & ResBlock 6 & ResBlock 8 & ResBlock 10 & ResBlock 12 \\
            \midrule
            RAVEN & 12.5 & \textbf{53.09} & 53.64 & 56.49 & 56.04 & 54.50 & 39.12 & 40.80 \\
            CVR & 25 & 47.56 & 58.73 & 67.56 & \textbf{74.65} & 79.20 & 75.42 & 76.22 \\
            SVRT & 50 & 51.96 & 51.84 & 53.49 & 66.33 & \textbf{87.33} & 88.92 & 90.22 \\
            Bongard-HOI & 50 & 55.51 & 56.46 & 55.93 & \textbf{61.97} & 61.23 & 60.48 & 61.02 \\
            Bongard-LOGO & 50 & 56.16 & 62.50 & 66.50 & \textbf{69.48} & 68.74 & 68.02 & 65.58 \\
        \end{tabular}
    }
    \label{tab:symbol}
\end{table*}

\begin{figure*}[!t]
    \centering
    \resizebox{1.0\textwidth}{!}{
    \includegraphics[width=\textwidth]{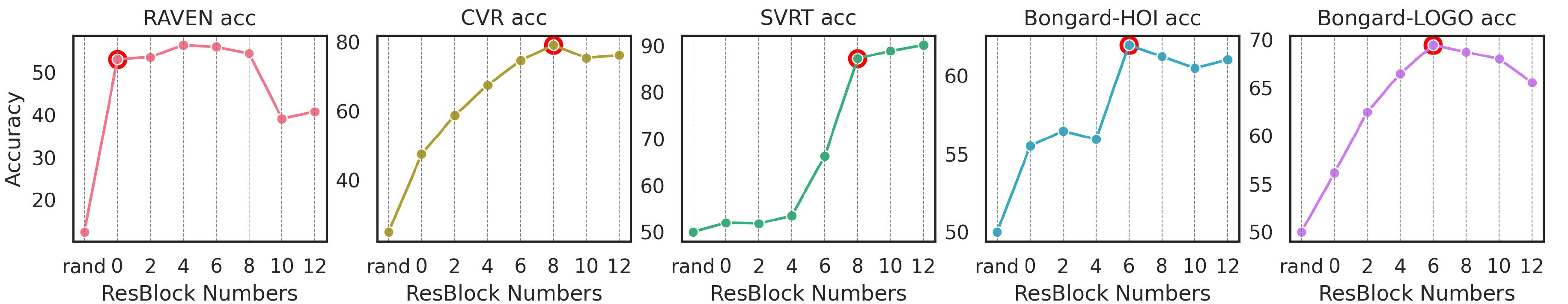}
    }
    \caption{Performance curve of varying encoder depth in symbolization stage. We present the results across RAVEN, CVR, SVRT, Bongard-LOGO, and Bongard-HOI. The highlight points refer to the distinct inflection points.}
    \label{fig:5}
\end{figure*}

\subsection{Optimal Symbolization Depth}

\begin{table*}[t]
    \centering
    \caption{Impact of shared reasoner implemented with different architectures in the Shared-Reason-Only type. 
    }
    \resizebox{\textwidth}{!}{
        \begin{tabular}{c|ccccccccc}
            & \textbf{RAVEN} & \textbf{CVR} & \textbf{SVRT} & \textbf{Bongard-HOI} & \textbf{Bongard-LOGO} & 
            \textbf{CoPhyBall} & \textbf{CoPhyBlocktower} & \textbf{CoPhyCollision} & \textbf{VQA} \\
            \midrule
            Both Separated (ad-hoc) & 53.40 & 78.30 & 92.50 & 66.43 & 72.30 & 6.53 & 0.39 & 2.91 & 54.08\\
            \midrule
            MLP & 55.72 & 72.40 & 86.64 & 57.65 & 69.64 & \textbf{8.01} & \textbf{0.84} & \textbf{2.98} & \textbf{50.06} \\
            CNN & \textbf{58.47} & 74.04 & 86.20 & 58.47 & 62.16 &  8.87 & 1.01 & 3.04 & 41.47 \\
            Transformer & 49.36 & 78.48 & 84.20 & \textbf{62.50} & 69.32 & 9.34 & 1.80 & 3.74 & 32.64\\
            GCN & 52.72 & \textbf{78.75} & \textbf{88.10} & 61.36 & \textbf{71.27} & 8.17 & 0.95 & 3.31 & 42.19\\
            Neuro-Symbolic & 27.89 & 69.69 & 77.72 & 55.21 & 59.68 & 11.23 & 3.04 & 5.31 & 36.62\\ 
        \end{tabular}
    }    
    \label{tab:baseline}
\end{table*}

\begin{table*}[t]
    \centering
    \caption{Comparison between SOTA baselines and our One-for-All model. 
    The ``-'' means the original papers do not provide the results on the corresponding dataset.
    }
    \resizebox{1.0\textwidth}{!}{
    \begin{tabular}{c|cccccc}
        & \textbf{RAVEN} & \textbf{CVR} & \textbf{SVRT} & \textbf{Bongard-HOI} & \textbf{Bongard-LOGO} & \textbf{VQA} \\
        \midrule
        \multirow{2}{*}{Ad-hoc SOTA (Lite)}
         & 53.4~\cite{zhang2019raven} & 78.3~\cite{zerroug2022benchmark} & 92.5~\cite{messina2022recurrent} & 
        57.7~\cite{shu2022test} &  65.4~\cite{nie2020bongard} & 54.1~\cite{Antol_2015_ICCV} \\
        & (ResNet)& (ResNet-50 SSL)& (ResNet-50) & (CoCoOp) & (ProtoNet) & (LSTM+CNN)\\
        One-for-All (MLP Reasoner) & 55.7 & 72.4 & 86.6 & 57.7 & 69.6 & 50.1 \\ \hline
        \multirow{2}{*}{Ad-hoc SOTA (Heavy)}
        & 94.1~\cite{spratley2020closer}  & \multirow{2}{*}{--} & 96.9~\cite{messina2022recurrent} &  66.4~\cite{raghuraman2023crossimage} & 75.3~\cite{raghuraman2023crossimage} & 76.4~\cite{jiang2020defense}  \\
        &  (Rel-AIR) &  & (CorNet-S) & (TPT) & (SVM+MIMIC) & (ROUGE\_L) \\
        One-for-All (MiniGPT-4 Reasoner) & 21.6 & 58.3 & 97.8 & 53.8 & 54.8 & 63.6 \\ 
    \end{tabular}}
    
    \label{tab:one-for-all}
\end{table*}

Next, we aim to identify the boundary between the two stages by probing the depth of the symbol encoder as shown in Fig.~\ref{fig:enter-label3}. 
The shared reasoner we use is an MLP.
By observing the changes of the \textit{accuracy}, we hope to find the distinct \textbf{inflection point} and termination of the symbolization grounding stage.
To ensure fairness, we employ the ResNet18~\cite{he2016deep} encoder across the following 2D datasets: RAVEN~\cite{zhang2019raven}, CVR~\cite{zerroug2022benchmark}, SVRT~\cite{li2011comparing}, Bongard-LOGO~\cite{nie2020bongard}, Bongard-HOI~\cite{jiang2022bongard}.
For each benchmark, we train individual models to their optimal performance and then interrupt the trained networks at various depths to probe the symbolization termination points.
We connect the output of the separated symbol encoders to a shared reasoner and record the accuracy at different interruption points, which is treated as evidence of symbolization termination.
The results are illustrated in Tab.~\ref{tab:symbol},
which presents outcomes of various datasets following the deepening of the symbolization encoder.
We also visualize the results in Fig \ref{fig:5}.

As observed in Fig.~\ref{fig:5} and Tab.~\ref{tab:symbol}, for each benchmark, the network depth exhibits an initial increase followed by a plateau. The inflection point positions vary across different tasks due to their varying levels of difficulty and the requisite degree of symbol abstraction. 
The inflection point of Bongard-HOI is much deeper than RAVEN, showing that the former is more difficult to symbolize and needs a deeper symbolization network to acquire a sophisticated high-dimensional feature.
These results verify the necessity of employing symbolization networks with varying depths for datasets of different complexities and illustrate the reasonable boundary between the two stages.

\subsection{One-for-All Reasoner Architecture}
Next, we aim to figure out the suitable architecture for the reasoner and test its effect in the Shared-Reason-Only design.
We choose 9 cross-domain datasets, RAVEN~\cite{zhang2019raven}, CVR~\cite{zerroug2022benchmark}, SVRT~\cite{li2011comparing}, Bongard-HOI~\cite{jiang2022bongard}, Bongard-LOGO~\cite{nie2020bongard}, Filtered-CoPhy~\cite{janny2022FilteredCoPhy}, VQAv2~\cite{Goyal_2017_CVPR} to conduct our experiment, as solving various reasoning problems with different domains can better demonstrate the model's reasoning abilities. 

We design task-specific encoders and heads based on the requirements of different tasks.
As for reasoner, we test CNN, MLP, Transformer~\cite{vaswani2017attention}, GCN~\cite{gori2005new}, hybrid Neuro-Symbolic model~\cite{Mao2019NeuroSymbolic}, and MiniGPT-4~\cite{zhu2023minigpt, chen2023minigptv2}. 
We first individually train each dataset to obtain the best results with the separated encoders and separated heads. 
Then we perform joint training on multiple datasets using one shared reasoner.

In Tab.~\ref{tab:baseline}, we conclude that
within all the architectures, MLP surprisingly performs the best on four datasets and is comparable in the other five datasets. Besides, the GCN performs well on three datasets following the previous experience in reasoning works~\cite{gori2005new}.
However, other architectures that are usually thought to be more advanced like Transformer do not show obvious advantages. 
Thus, we choose the MLP as a lite reasoner in our One-for-All model.

In Tab.~\ref{tab:one-for-all}, our One-for-All model demonstrates impressive performance even compared to the ad-hoc state-of-the-art (SOTA) across a majority of tasks. 
We divide the SOTA according to their complexity into lite and heavy, as depicted in Tab.~\ref{tab:one-for-all}.
From the result, One-for-All is comparable with the lite ad-hoc SOTA and even outperforms the lite SOTA on some datasets like RAVEN.
This experiment demonstrates that the reasoning stage has quite a different performance-parameter relationship with recognition tasks. 
A lite reasoner may also perform well on reasoning if trained on multi-domain tasks.

Since reasoning ability cannot be solely measured by accuracy, we also evaluate reasoning capabilities using \textit{reasoning consistency}. For each task, we use the same encoder and reasoner parameters with two questioning methods: \textit{``What is the answer to this question?''} and \textit{``Is a certain option correct?''}. A model with good reasoning ability should yield consistent results across both methods, unlike a random model which may be inconsistent. We use the F1 score to measure the consistency between these methods, as shown in Tab.~\ref{tab:consistency}. Our One-for-All model, trained jointly on multiple datasets, shows higher consistency compared to models trained individually, demonstrating its potential in genuine reasoning.

\begin{figure}[t]
    \centering
    \includegraphics[width=1.0\textwidth]{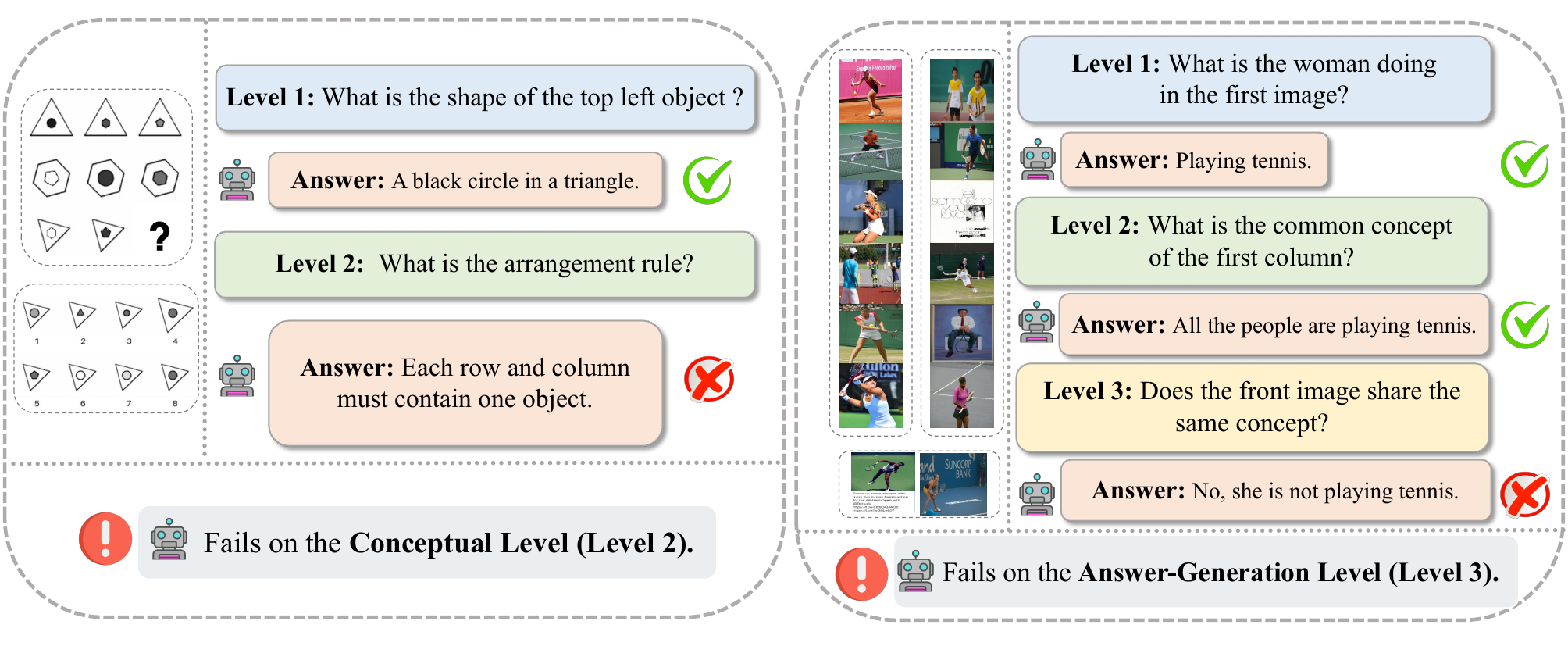}
    \caption{Failure case analysis of LLM-based model. On RAVEN, it fails on the Symbolization level; while on Bongard-HOI, it fails on the Answer Generation level.}
    \label{fig:LLM}
\end{figure}

\begin{table*}[!t]
    \centering
    \caption{Comparison of accuracy and reasoning consistency between the ad-hoc methods and our one-for-all model. Our one-for-all model demonstrates decent performance.}
    \label{tab:consistency}
    \resizebox{0.65\textwidth}{!}{
        \begin{tabular}{c|c|cccc}
         \textbf{Metric} & \textbf{Model} & \textbf{RAVEN} & \textbf{CVR} & \textbf{SVRT} & \textbf{Bongard-HOI} \\
    \midrule
        \multirow{2}{*}{Accuracy} & Both-Separated & 53.40 & \textbf{74.04} & 86.20 & \textbf{61.64} \\
         & One-for-All & \textbf{55.72} & 72.40 & \textbf{86.64} & 57.65 \\
    \midrule
        \multirow{2}{*}{Consistency} & Both-Separated & 76.94 & 52.35 & 99.42 & 98.93 \\
         & One-for-All & \textbf{77.89} & \textbf{53.51} & \textbf{99.73} & \textbf{99.55} \\ 
    \end{tabular}
    }

\end{table*}

To further evaluate the performance of LLM, we adopt MiniGPT-4~\cite{zhu2023minigpt} as the shared reasoner. 
Our One-for-All model also shows superiority given a similar model scale.
Surprisingly, our lite One-for-All reasoner surpasses the MiniGPT-4 on specific tasks, \eg, RAVEN, and Bongard-HOI, providing compelling evidence that there is no absolute positive correlation between the number of model parameters and the model's reasoning capabilities. 

To analyze the performance of LLM-based models,
we probe the task according to our two-stage framework design and examine them separately:
(1) \textit{Symbolization}: whether LLM-based models can recognize the elements of the problem.
(2) \textit{Conceptual}: whether LLM-based models can learn specific concepts behind the tasks and reason about them.
(3) \textit{Answer Generation}: whether LLM-based models can utilize the concepts it learns to solve problems.
Using MiniGPT-4~\cite{zhu2023minigpt} as a representative,
we summarize the typical responses of LLM-based models to three-level problems in RAVEN and Bongard in Fig.~\ref{fig:LLM}.

We find that LLMs may encounter certain hallucination circumstances~\cite{xu2024hallucination,yao2023llm,ji2023towards} while solving visual reasoning tasks. As shown in Fig.~\ref{fig:LLM}, for the RAVEN problem, MiniGPT-4 succeeds in the first level of identifying the object while failing in the second stage of reasoning with the arrangement rule.
For RAVEN problems, MiniGPT-4 fails to accurately identify the logical patterns.
For the Bongard problem, MiniGPT-4 succeeds in the first level of recognizing human activity and the second level of grasping reasoning logically, yet it fails at the answer generation level and gets lost when utilizing rules to answer 
questions.
Given the above cases, we can gain an understanding of the shortcomings of the LLM-based models in reasoning tasks, namely its good concept comprehension ability but insufficient performance in logical reasoning and answer generation. 
In future work, we plan to train our framework with a larger LLM on more visual reasoning datasets in different domains.
We believe our work would inspire both visual and general reasoning studies.

\subsection{Approximation Principle Verification}
Next, we verify that training the reasoner with data from multiple domains can guarantee a reasoner with better generalization ability.
We conduct experiments on SVRT~\cite{li2011comparing}, Bongard-HOI~\cite{jiang2022bongard}, Balls task from Filtered-CoPhy~\cite{janny2022FilteredCoPhy}, Collision task from Filtered-CoPhy~\cite{janny2022FilteredCoPhy}, and VQAv2~\cite{Goyal_2017_CVPR}. 
These datasets encompass 2D puzzles, 3D video, and VQA tasks, offering diverse and multi-modal data. 
We utilize the Filtered-CoPhy Collision task as the benchmark for testing.

\begin{wraptable}{r}{0.56\textwidth}
    \centering
    \caption{Validation of ``approximation principle'' using the Collision task in Filtered-CoPhy. We train the reasoner using data from more and more domains and then transfer the reasoner to the target Filtered-CoPhy in inference. We use the keypoint error as the metric of Filtered-CoPhy (the smaller the better). We set two tracks, \ie, training with 1,000 or 3,000 samples from each training dataset.
    }
    \resizebox{0.52\textwidth}{!}{
        \begin{tabular}{c|cccc|cc}
            & SVRT & Bongard-HOI & CoPhyBall & VQAv2 & 1,000 Samples$\downarrow$ & 3,000 Samples$\downarrow$ \\
            \midrule
            \multirow{4}{*}{$n$ = 1} & \checkmark &  &  &  &  13.75 & 13.09 \\
             &  &\checkmark  &  &  &  14.40 & 9.90 \\
             &  &  & \checkmark &  &  18.04 & 24.80 \\
             &  &  &  & \checkmark &  14.00 & 14.85\\
            \midrule
            \multirow{6}{*}{$n$ = 2} & \checkmark & \checkmark &  &  &  11.34 & 12.76 \\
             & \checkmark &  & \checkmark &  &  10.13  & 7.41\\
             & \checkmark &  &  & \checkmark &  11.24 & 10.13\\
             &  & \checkmark &  \checkmark&  &  9.99 & 9.53\\
             &  & \checkmark &  & \checkmark &  13.72 & 5.56\\
             &  &  & \checkmark & \checkmark &  9.34 & 6.29\\
            \midrule
            \multirow{4}{*}{$n$ = 3} & \checkmark & \checkmark & \checkmark &  &  8.06 & 7.39\\
             & \checkmark & \checkmark &  & \checkmark &  8.61 & 6.75\\
             & \checkmark &  & \checkmark & \checkmark &  \textbf{5.54} & 6.46\\
             &  & \checkmark & \checkmark &  \checkmark &  10.28 & \underline{6.18} \\
            \midrule
            $n$ = 4 & \checkmark & \checkmark & \checkmark & \checkmark &  \underline{5.85} & \textbf{4.88}\\
        \end{tabular}
    }
    \label{tab:generalize}
\end{wraptable}

We involve more and more cross-domain datasets to train the reasoner and pair it with the separated encoder of the target test dataset.
We report the comparison in Tab.~\ref{tab:generalize}.
Given the inherent gaps between the various datasets, we introduce a highly lightweight, MLP-based adapter into the reasoner prior. 
To equalize the contribution of each dataset to the reasoning engine, we adjust the sample sizes used for training across datasets. Specifically, we use sample sizes of 1,000 and 3,000. 

Tab.~\ref{tab:generalize} reveals a gradual improvement of the reasoner with an increasing number of training datasets. 
Although handling more datasets from diverse domains significantly enhances the complexity, the trained reasoner performs well on the out-of-domain Filtered-CoPhy. 
This shows that the reasoner would concentrate on task-agnostic pure reasoning as the domains of the training dataset increase, which validates our ``approximation principle''.

\subsection{Additional Ablation Study}
This section mainly involves experiments in pretraining the encoder using the large-scale image dataset and using the CLIP model as a general encoder. 
For more experiments please refer to the supplementary material.

\begin{table}[t]
  \centering
  \caption{Symbol encoders trained from scratch and pre-trained on ImageNet. }
  \resizebox{0.5\textwidth}{!}{
  \begin{tabular}{c|ccc}
     & \textbf{RAVEN} & \textbf{CVR} & \textbf{SVRT}  \\
    \midrule
    Trained from Scratch & \textbf{53.56} & 72.40 & \textbf{87.84} \\
    ImageNet Pretrained & 52.21 & \textbf{74.82} & 85.53 \\
  \end{tabular}}
  \label{tab:pretrain}
\end{table}
\begin{table}[t]
    \centering
    \caption{Comparison between CLIP as a general encoder and our best One-for-All model. Our One-for-All model surpasses using CLIP as a general encoder in most tasks.  
    }
    \resizebox{0.9\textwidth}{!}{
    \begin{tabular}{c|ccccc}
        & \textbf{RAVEN} & \textbf{CVR} & \textbf{SVRT} & \textbf{Bongard-HOI} & \textbf{Bongard-LOGO} \\
        \midrule
        One-for-All (shared reasoner) & \textbf{55.72} & 72.40 & \textbf{86.64} & \textbf{57.65} & \textbf{69.64}\\
        CLIP (both shared) & 23.59 & \textbf{74.60} & 65.64 & 51.10 & 58.09\\
        
    \end{tabular}
    }
    \label{tab:CLIP}
\end{table}

{\bf Pre-trained Model.}
We present the ablation on using the pre-trained model or not for symbol encoder in Tab.~\ref{tab:pretrain}. 
We select RAVEN\cite{zhang2019raven}, CVR\cite{zerroug2022benchmark} and SVRT\cite{li2011comparing} dataset, and we employ ImageNet\cite{deng2009imagenet} as the pretrain dataset. 
We can find that the results are very close. The possible reason is that there is an obvious domain gap between ImageNet and the three reasoning datasets.

{\bf CLIP as General Encoder.}
We test whether CLIP~\cite{radford2021learning}, a general and large foundation model can do well acting as the general symbol encoder.
We utilize CLIP as the visual encoder for multi-modal datasets, followed by an MLP as the reasoner, and employ task-specific head networks.
As shown in Tab.~\ref{tab:CLIP}, we find that using CLIP yields inferior outcomes to the best One-for-All approach, even after fine-tuning.
This validates that even large models such as CLIP are unable to accomplish the symbolization of different datasets, thereby affirming the rationale behind our separated encoder, shared reasoner framework design.
\section{Conclusion}
In this work, we conclude a two-stage perspective for visual reasoning: the symbolization transforms data into symbolic representation, while the reasoning performs logical reasoning. 
We show that compared with symbolization, reasoning is more task-agnostic and can be shared by cross-domain tasks.
Thus we introduce a concise framework consisting of separated symbol encoders and a shared reasoner.
It is crucial to select the complexity of the symbolization and use multi-domain data to train the reasoner to pursue generalization.
Our framework achieves decent performance on multiple datasets with domain gaps. 
We believe our work would pave the way for generalizable visual reasoning systems.

\section*{Acknowledgments}
This work is supported in part by the
National Natural Science Foundation of China under Grants No.62306175.

\bibliographystyle{splncs04}
\bibliography{main}

\newpage

{\centering
        \Large
        \textbf{Take A Step Back: Rethinking the Two Stages in Visual Reasoning}\\
        \vspace{0.5em}Supplementary Material \\
        \vspace{1.0em}}

In this supplementary, we provide the additional contents as follows:

Sec.~\ref{sec:dataset}: Dataset Details.

Sec.~\ref{sec:experiment}: Experimental Details.

Sec.~\ref{sec:ablation}: Additional Ablation Study.

\section{Dataset Details}
\label{sec:dataset}
{\bf RAVEN}~\cite{zhang2019raven} is a 2D puzzle benchmark for visual reasoning.
It contains 42,000 problems in the train set, 14,000 problems in the validation set, and 14,000 problems in the test set.
Each question contains a $3 \times 3$ grid arrangement of images, where the bottom right corner is empty, and 8 options.
The task is to choose the option that best matches the original image's shape, size, position relationship, and other geometric features from the 8 options to fill in the blank.

{\bf CVR}~\cite{zerroug2022benchmark} is a 2D puzzle benchmark for visual reasoning.
To control training difficulty, we choose the elementary tasks which focus on one single object attribute like color, shape, positional attributes, \etc.
For each subset task, we generate 1,000 problems in the train set, 1,000 problems in the validation set, and 1,000 problems in the test set.
Each question contains 4 images with different object attributes.
The task is to identify the outlier based on the "odd-one-out" rule, where one item possesses an attribute different from that shared by the others.

{\bf SVRT}~\cite{li2011comparing} is a 2D puzzle benchmark for visual reasoning.
It contains 200,000 problems in the train set, 50,000 problems in the validation set, and 50,000 problems in the test set.
Specifically, to control difficulty, we choose the \#21 problems as our benchmarks, which aim to decide whether the two shapes in the image are the same or different.

{\bf Bongard-HOI}~\cite{jiang2022bongard} is a typical 2D visual reasoning dataset, which focuses on human-object interaction (HOI) and is constructed using the HAKE~\cite{DBLP:journals/pami/LiLWLQXXFL23} dataset by extracting few-shot instances to form concepts. 
The Bongard-HOI dataset contains 21,956 few-shot instances of the HAKE dataset in the train set, which share 118 different visual concepts; 
17,184 few-shot instances in the validation set, which share 167 different visual concepts; 13,941 few-shot instances in the test set, which share 166 different visual concepts. 
The question samples in Bongard-HOI can be described as follows: six images that all share a common concept \textit{C} are given, forming a positive sample set \textit{P}.
Another six images that do not have the concept \textit{C} are given, forming a negative sample set \textit{N}. 
For each question, a test sample query is constructed from two images, and the model needs to determine whether the two images in the query have the concept \textit{C} shared by the positive sample set.
The visual concept in the Bongard-HOI dataset may be: whether riding a bicycle or whether using a laptop. It involves reasoning about human and object interaction behavior in the real world.

{\bf Bongard-LOGO}~\cite{nie2020bongard} is a typical 2D visual reasoning dataset. It is a modification of the original Bongard Problem, focusing on few-shot classification instead of learning concepts from natural language descriptions. The Bongard-LOGO problem can be described as follows: given six images that share a common concept, an equal number of images that do not share that concept, and two query images, determine whether the two query images share the concept depicted in the first six images. 
The Bongard-LOGO dataset comprises 12,000 images and primarily involves three categories of problems: the Free-Form shape problem, which includes 3,600 samples and pertains to understanding randomly generated shapes; the Basic shape problem, which includes 4,000 samples and revolves around concepts formed by combining one or two out of 627 different shapes designed by humans; and the Abstract shape problem, which includes 4,400 samples and utilizes shapes from the Basic shape problem but involves more complex attributes and combinations.

{\bf Filtered-CoPhy}~\cite{janny2022FilteredCoPhy} is a 3D physical visual reasoning dataset that requires models to learn causal relationships, physical parameters, and physical laws through training. 
Each sample in the Filtered-CoPhy dataset contains the following: an initial state $A$ (a single image) and subsequent state $B$ (a video), a state $C$ modified by a $do$-$operator$ from $A$, and a subsequent video $D$ predicted from the initial state $C$. 
The task is to observe $A$, $B$, and $C$ and predict the situation in $D$. The Filtered-CoPhy dataset includes three tasks: BlockTower task, Collision task, and Balls task. 
The BlockTower task requires predicting the stability and motion of stacked square blocks and includes 146k samples. 
The Collision task requires predicting the collision process and subsequent motion states of objects with different shapes and includes 50k samples. 
The Balls task requires predicting the motion and collision of multiple balls and includes 100k samples.
In this experiment, since we do not need to reconstruct the video, the metric we use is the MSE of predicted keypoints and real keypoints scaled by 200.

{\bf VQAv2}~\cite{Goyal_2017_CVPR} is a typical VQA benchmark specifically designed for 2D visual reasoning. 
It builds upon the basic VQA dataset but incorporates special modifications to achieve a balance between natural language questions and answers. 
The dataset utilizes images from the COCO~\cite{lin2014microsoft} dataset, which consists of 204k images depicting everyday human life, along with 614k natural language questions and 6 million natural language descriptions as answers. 
Each image in the VQAv2 dataset is accompanied by three natural language questions, each with ten options. The options contain repeated items, and our model is required to select the most suitable answer among these options. 
The balance of the VQAv2 dataset is reflected in the fact that for each question, there is a corresponding pair of images that are asked the same question, yet they generate different answers. 
This balance ensures that the model cannot solely rely on biases in the questions to obtain answers directly from natural language but must analyze the images to respond appropriately.

\section{Experimental Details}
\label{sec:experiment}
\subsection{Entanglement v.s. Disentanglement Analysis}
To verify our visual reasoning framework, 
we conduct separate tests on various datasets, utilizing the same encoder and reasoner configurations, and compare the performance of these two design paradigms across diverse datasets. 

In our experiments, we
train the models using five datasets: RAVEN~\cite{zhang2019raven}, CVR~\cite{zerroug2022benchmark}, SVRT~\cite{li2011comparing}, Bongard-HOI~\cite{jiang2022bongard}, and Bongard-LOGO~\cite{nie2020bongard}.
In Type-1 \textbf{Both-Separated}, the symbol encoder and logical reasoner are both separated and trained on each dataset respectively.
In Type-2 \textbf{Both-Shared}, the symbol encoder and reasoner are both shared and trained together with different datasets. Before the encoder is a tiny data-preprocessing to align dimensions.
In Type-3 \textbf{Shared-Encoder-Only}, the data are initially processed through a shared encoder, then the outputs are fed into the separated reasoners and heads to yield the final results. 
In Type-4 \textbf{Shared-Reasoner-Only}, the data from each dataset are individually processed through its unique encoder, and then the outputs are combined and input into a shared reasoner.

To control variables and facilitate model sharing, for all types above we adopt ResNet-18~\cite{he2016deep} as the encoder and an MLP as the reasoner. 
During the training process, we finetune the hyperparameters of each Adam optimizer~\cite{kingma2014adam} for each separate part of the network for each dataset. For the shared networks, we set the learning rate 5e-5 and weight decay 1e-7.
Due to the varying complexities of different datasets, the required number of training epochs also varies, so we control the entry timing of joint training to ensure that all datasets reach their best results simultaneously.
Relative hyperparameters are shown in Tab.~\ref{tab:shared}.

\begin{table}[t]
    \centering
    \caption{Hyperparameters of entanglement v.s. disentanglement experiment. We showcase hyperparameters for training shared-encoder, shared-reasoner, both-shared, and both-separated. ``Entry Epoch'' refers to the epoch at which a corresponding dataset enters the training process,
    with 0 indicating initial entry and other numerical values indicating delayed entry for specific epochs. 
    }
    \resizebox{0.8\textwidth}{!}{
    \begin{tabular}{c|cccc}
        & \textbf{Batch Size} & \textbf{Learning Rate} & \textbf{Weight Decay} & \textbf{Entry Epoch} \\
        \midrule
        RAVEN & 16 & $7.18 \times 10^{-5}$ & $1.00 \times 10^{-8}$ & 42 \\
        CVR & 8 & $2.99 \times 10^{-4}$ & $1.40 \times 10^{-7}$ & 63 \\
        SVRT & 8 & $7.42 \times 10^{-5}$ & $6.66 \times 10^{-5}$ & 0 \\
        Bongard-HOI & 16 & $1.21 \times 10^{-4}$ & $1.84 \times 10^{-5}$ & 30\\
        Bongard-LOGO & 16 & $3.12 \times 10^{-5}$ & $1.58 \times 10^{-5}$ & 0\\
    \end{tabular}
    }
    \label{tab:shared}
\end{table}

\subsection{Optimal Symbolization Depth}
In our two-stage framework, the symbolization stage is intermediate between multimodal input data and the abstract logical reasoning stage. 
This process involves symbolizing individual input signals on a latent space capable of performing reasoning.

We employ ResNet-18~\cite{he2016deep} as the encoder and MLP as the reasoner to probe the depths of symbolization. 
Initially, we train the encoder-reasoner framework separately on each dataset and then interrupt the trained encoders at various depths to probe the symbolization termination points.
We extract the interrupted encoders, freeze them, connect the outputs of the separated symbol encoders to a shared reasoner, and record the accuracy at different interruption points, which is treated as evidence of symbolization termination.
We record the symbolization depth and accuracy curves for different datasets.
For the shared reasoner, we use the Adam optimizer~\cite{kingma2014adam} with the learning rate 5e-5 and weight decay 1e-7.

To verify the importance of obtaining the most suitable symbolization depth for the overall task, we conduct supplementary experiments.
We extract the symbol encoders with the inflection point depth achieved from each experiment and connect them to a shared reasoner MLP.
We train the reasoner and compare their final scores with those of the One-for-All reasoner in full depth. 
The results in Tab.~\ref{tab:depth} show that the network with the inflection point depth outperforms the ordinary One-for-All network.
This confirms the significance of selecting the most suitable depth for different datasets and validates the importance of probing symbolization depths for different datasets.

\begin{table}[t]
    \centering
    \caption{Comparison between encoders with different depths. We compare the result of all tasks sharing the same full encoder depth with the result of the encoder whose depth of each task is the depth that the inflection point occurs.
    }
    \resizebox{0.9\textwidth}{!}{
    \begin{tabular}{c|ccccc}
        & \textbf{RAVEN} & \textbf{CVR} & \textbf{SVRT} & \textbf{Bongard-HOI} & \textbf{Bongard-LOGO} \\
        \midrule
        One-for-All (Full Depth) & 55.72 & \textbf{72.40} & 86.64 & 57.65 & 69.64 \\
        One-for-All (Inflection Point Depth) & \textbf{56.47} & 68.01 & \textbf{87.21} & \textbf{57.91} & \textbf{69.91}\\
        
    \end{tabular}
    }
    \label{tab:depth}
\end{table}

\begin{table*}[!th]
    \centering
    \caption{Depict of the hyperparameters we choose to train the One-for-All model. We give batch size, learning rate, weight decay, and entry epoch we choose to train MLP, CNN, Transformer, GCN, and Neural Symbolic network.
    The hyperparameter ``Entry Epoch'' refers to the epoch at which a corresponding dataset enters the training process,
    with 0 indicating initial entry and other numerical values indicating delayed entry for specific epochs. 
    The parameter is set due to variations in task difficulties, as certain tasks can achieve optimal performance without undergoing a full 100 epochs of training.
    }
    \resizebox{0.8\textwidth}{!}{
    \begin{tabular}{c|c|cccc}
        \textbf{Baselines} & \textbf{Benchmarks} & \textbf{Batch Size} & \textbf{Learning Rate} & \textbf{Weight Decay} & \textbf{Entry Epoch} \\
        \midrule
        \multirow{9}{*}{MLP} & RAVEN & 16 & $7.18 \times 10^{-5}$ & $1.00 \times 10^{-8}$ & 42 \\
        & CVR & 8 & $2.99 \times 10^{-4}$ & $1.40 \times 10^{-7}$ & 63 \\
        & SVRT & 8 & $7.42 \times 10^{-5}$ & $6.66 \times 10^{-5}$ & 0 \\
        & Bongard-HOI & 16 & $1.21 \times 10^{-4}$ & $1.84 \times 10^{-5}$ & 30\\
        & Bongard-LOGO & 16 & $3.12 \times 10^{-5}$ & $1.58 \times 10^{-5}$ & 0\\
        & CoPhyBalls & 16 & $9.36 \times 10^{-4}$ & $1.98 \times 10^{-5}$ & 0\\
        & CoPhyTower & 16 & $1.99 \times 10^{-3}$ & $3.23 \times 10^{-7}$ & 0\\
        & CoPhyCollision & 16 & $1.10 \times 10^{-3}$ & $2.80 \times 10^{-8}$ & 0\\
        & VQA & 16 & $6.82 \times 10^{-3}$ & $1.21 \times 10^{-6}$ & 0\\
        \midrule
        \multirow{9}{*}{CNN} & RAVEN & 16 & $1.32 \times 10^{-5}$ & $1.82 \times 10^{-8}$ &  42\\
        & CVR & 8 & $2.11 \times 10^{-4}$ & $1.28 \times 10^{-4}$ & 63 \\
        & SVRT & 8 & $1.21 \times 10^{-4}$ & $1.04 \times 10^{-6}$ &  0\\
        & Bongard-HOI & 16 & $1.24 \times 10^{-4}$ & $1.34 \times 10^{-6}$ & 30\\
        & Bongard-LOGO & 16 & $2.64 \times 10^{-5}$ & $1.33 \times 10^{-8}$ &0 \\
        & CoPhyBalls & 16 & $5.00 \times 10^{-5}$ & $1.00 \times 10^{-7}$ & 0\\
        & CoPhyTower & 16 & $5.00 \times 10^{-5}$ & $1.00 \times 10^{-7}$ & 0\\
        & CoPhyCollision & 16 & $5.00 \times 10^{-5}$ & $1.00 \times 10^{-7}$ & 0\\
        & VQA & 16 & $1.18 \times 10^{-4}$ & $2.87 \times 10^{-4}$ & 0\\
        \midrule
        \multirow{9}{*}{Transformer} & RAVEN & 16 & $2.65 \times 10^{-5}$ & $1.96 \times 10^{-4}$ & 42 \\
        & CVR & 8 & $1.22 \times 10^{-4}$ & $1.08 \times 10^{-4}$ &  63\\
        & SVRT & 8 & $4.20 \times 10^{-5}$ & $4.35 \times 10^{-5}$ &  0\\
        & Bongard-HOI & 16 & $1.85 \times 10^{-6}$ & $7.97 \times 10^{-6}$ & 30\\
        & Bongard-LOGO & 16 & $1.01 \times 10^{-5}$ & $2.79 \times 10^{-6}$ &0 \\
        & CoPhyBalls & 16 & $4.34 \times 10^{-4}$ & $3.78 \times 10^{-6}$ & 0\\
        & CoPhyTower & 16 & $2.58 \times 10^{-5}$ & $2.48 \times 10^{-6}$ & 0\\
        & CoPhyCollision & 16 & $5.46 \times 10^{-5}$ & $7.01 \times 10^{-8}$ & 0\\
        & VQA & 16 & $3.06 \times 10^{-5}$ & $3.82 \times 10^{-5}$ & 0\\
        \midrule
        \multirow{9}{*}{GCN} & RAVEN & 16 & $5.95 \times 10^{-5}$ & $6.54 \times 10^{-7}$ & 42 \\
        & CVR & 8 & $4.26 \times 10^{-4}$ & $1.18 \times 10^{-5}$ & 63 \\
        & SVRT & 8 & $2.63 \times 10^{-4}$ & $1.70 \times 10^{-6}$ & 0 \\
        & Bongard-HOI & 16 & $9.30 \times 10^{-5}$ & $3.11 \times 10^{-5}$ & 30\\
        & Bongard-LOGO & 16 & $3.62 \times 10^{-5}$ & $1.87 \times 10^{-8}$ & 0\\
        & CoPhyBalls & 16 & $2.26 \times 10^{-3}$ & $3.39 \times 10^{-8}$ & 0\\
        & CoPhyTower & 16 & $2.33 \times 10^{-3}$ & $1.08 \times 10^{-8}$ & 0\\
        & CoPhyCollision & 16 & $2.77 \times 10^{-3}$ & $1.01 \times 10^{-8}$ & 0\\
        & VQA & 16 & $7.03 \times 10^{-5}$ & $9.31 \times 10^{-7}$ & 0\\
        \midrule
        \multirow{9}{*}{Neuro Symbolic} & RAVEN & 16 & $3.38 \times 10^{-5}$ & $8.56 \times 10^{-7}$ & 42 \\
        & CVR & 8 & $1.81 \times 10^{-4}$ & $2.69 \times 10^{-8}$ & 63 \\
        & SVRT & 8 & $3.14 \times 10^{-5}$ & $3.52 \times 10^{-8}$ & 0 \\
        & Bongard-HOI & 16 & $1.85 \times 10^{-6}$ & $7.97 \times 10^{-6}$ & 30\\
        & Bongard-LOGO & 16 & $1.16 \times 10^{-5}$ & $1.34 \times 10^{-4}$ & 0\\
        & CoPhyBalls & 16 & $1.08 \times 10^{-5}$ & $4.81 \times 10^{-4}$ & 0\\
        & CoPhyTower & 16 & $3.58 \times 10^{-5}$ & $2.50 \times 10^{-5}$ & 0\\
        & CoPhyCollision & 16 & $2.83 \times 10^{-3}$ & $3.33 \times 10^{-6}$ & 0\\
        & VQA & 16 & $5.07 \times 10^{-3}$ & $1.99 \times 10^{-4}$ & 0\\
        
    \end{tabular}
    }
    \label{tab:one-for-all}
\end{table*}

\begin{table}[!th]
    \centering
    \caption{The prompting questions we use when we finetune and test the performance of MiniGPT-4. For each dataset, we design a specific prompt that enables the LLM to tackle the problem. 
    }
    \resizebox{0.95\textwidth}{!}{
    \begin{tabular}{c|p{0.85\textwidth}}
        \textbf{Datasets} & \multicolumn{1}{c}{\centering \textbf{Prompts}} \\
        \midrule
        \multirow{17}{*}{RAVEN} & \textit{I have a classic Raven's problem for you to solve. In this problem, I will describe a series of patterns and ask you to identify the underlying pattern or rule. You will need to use your pattern recognition skills to infer the missing elements and provide the correct answer. Please pay attention to the relationships and transformations within the patterns. Here is an example. Please learn the rules from it. In the matrix, on the first row is a black circle in a large triangle, a dark hexagon in a large triangle, and a gray pentagon in a large triangle. On the second row is a white pentagon in a large hexagon, a black circle in a large hexagon, and a dark hexagon in a large hexagon. On the third row is a white hexagon in a large triangle, a black pentagon in a large triangle, with the last one missing. For the inner shape, according to the color rule, it should be gray. For each row, there is a circle, a hexagon, and a pentagon. So the inner shape should be a circle. Outside each row involves three same shapes. So the outside shape should be a large triangle. Now here comes the question on this image on the left. The answer would be one of the images on the right. Give me the index of your choice. Notice that your answer should be like "The answer is 7."}\\
        \midrule
        \multirow{3}{*}{CVR} & \textit{Take a look at these 4 images. Behind them, there are attributes such as shape, size, etc. One photo stands out distinctly from the others. Which one is the outlier? Please provide the corresponding index.}\\
        \midrule
        \multirow{1}{*}{SVRT} & \textit{Observe this image. Do the two geometric objects share the same shape?}\\
        \midrule
        \multirow{12}{*}{Bongard-HOI} & \textit{I have a classic Bongard-HOI Problem for you to solve. Here is an example of it. Please learn the rules from it. On the first row are six images that share the same concept of human activity; while the six images on the second row do not obey this concept. Now here are two question images. One of them is obeying the concept, and the other is not. Please tell me which one is obeying the concept, the latter one or the front? Your answer should be: The same concept is hitting the tennis ball. So the answer is the front one. Here is another question to check whether you understand this rule. On the first row are six images that share the same concept of human activity; while the six images on the second row do not obey this concept. On the third row are two question images. Please tell me which one obeys the concept, the front one or the latter?}\\
        \midrule
        \multirow{5}{*}{Bongard-LOGO} & \textit{Analyze this Bongard-LOGO problem. The images in the first row share a common concept, while the images in the second row lack that concept. I assert that the first image in the third row possesses that concept, whereas the second image in the third row lacks it. Am I correct? Please respond with a simple ``yes" or ``no".}\\
        \midrule
        \multirow{3}{*}{VQAv2} & \textit{Consider this image. It includes a photo accompanied by a question pertaining to the photo. Please select the best answer option, which should be a single number ranging from 0 to 9.}\\
        
    \end{tabular}
    }
    \label{tab:prompt}
\end{table}

\subsection{One-for-All Reasoner Architecture}
In our One-for-All reasoner experiment, we employ various reasoner architectures to observe the results. 
The reasoners utilized in the experiment include Multilayer Perceptron (MLP), Convolutional Neural Networks (CNN), Transformer~\cite{vaswani2017attention}, Graph Convolutional Networks (GCN)~\cite{gori2005new} and Neuro-Symbolic Approach~\cite{Mao2019NeuroSymbolic}.
For encoders, 2D puzzles like RAVEN~\cite{zhang2019raven}, CVR~\cite{zerroug2022benchmark}, SVRT~\cite{li2011comparing}, Bongard-HOI~\cite{jiang2022bongard}, Bongard-LOGO~\cite{nie2020bongard} utilize ResNet-18~\cite{he2016deep}, tasks of Filtered-CoPhy~\cite{janny2022FilteredCoPhy} use Graph Convolutional Networks (GCN)~\cite{gori2005new} and VQAv2~\cite{Goyal_2017_CVPR} adopts VGG-19~\cite{simonyan2014very} as the image encoder and LSTM~\cite{shi2015convolutional} as the text encoder.
Then, task-oriented MLP heads are connected after the reasoner to complete the classification and regression.
We present the training hyperparameters for all networks employed in the experiment in Tab~\ref{tab:one-for-all}.

Within the main text, we employ MiniGPT-4~\cite{zhu2023minigpt, chen2023minigptv2} as the reasoner. 
Since MiniGPT-4 can only accept one image and a piece of text prompt as input, we encode the several images into one whole to feed into the model.
For the VQAv2 dataset, both the images and questions are consolidated into a single image.
Further experimentation reveals that the effectiveness of MiniGPT-4 is heavily influenced by the results of fine-tuning and the employed natural language prompts.
The prompts used for each dataset are shown in Tab.~\ref{tab:prompt}.

\subsection{Approximation Principle Verification}
We conduct experiments across five datasets: SVRT~\cite{li2011comparing}, Bongard-HOI~\cite{jiang2022bongard}, Balls task from Filtered-CoPhy~\cite{janny2022FilteredCoPhy}, Collision task from Filtered-CoPhy~\cite{janny2022FilteredCoPhy}, and VQAv2~\cite{Goyal_2017_CVPR}. 
These datasets encompass 2D puzzles, 3D video, and VQA tasks, offering diverse and multi-modal data samples for the reasoner.
We utilize the Collision task from Filtered-CoPhy as the benchmark for testing.

We first train the encoder-reasoner framework on different numbers of training datasets, and then extract the reasoner and freeze it to combine it with the frozen encoder trained on the test dataset.
To fit the bias between the encoder and the reasoner, we use a lite fully connected layer of 512 neutrons as an adapter. 
During the finetuning process, we set a learning rate of 5e-5 and a weight decay of 1e-7.

\section{Additional Ablation Studies}
\label{sec:ablation}
In this section, we supplement the main text with an additional ablation study, focusing primarily on conducting experiments on adjusting hyperparameters and analyzing the reasoning performance of the two-stage framework. Specifically, we analyze the situations where the two-stage framework exhibits reasoning errors and provide visualizations to elucidate the limitations of the current baseline reasoning performance.

\subsection{Hyperparameters} 
\begin{figure}[!t]
    \centering
    \includegraphics[width=0.95\textwidth]{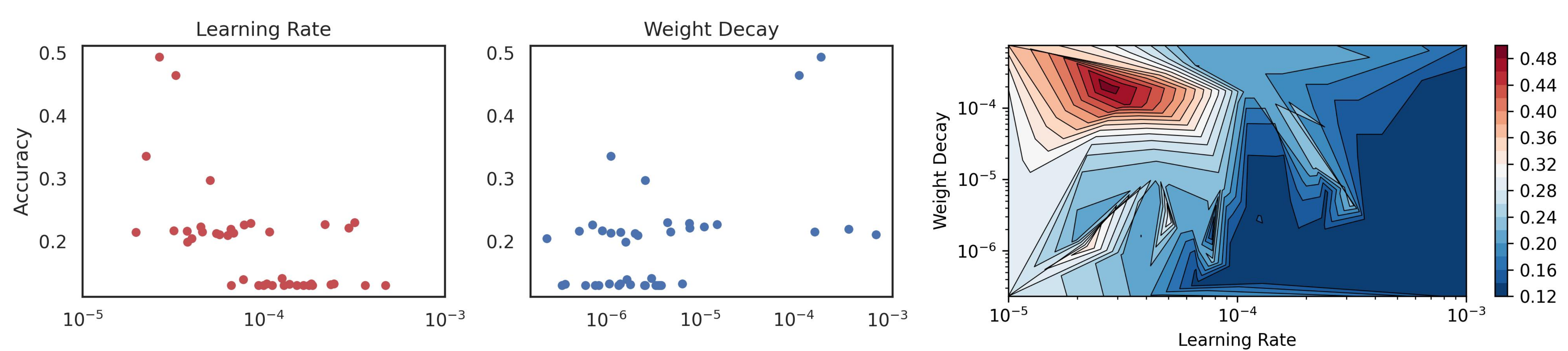}
    \caption{The hyperparameter finetuning process on the validation set. For each model in our experiment, we use Optuna to help us find the best hyperparameter value of the learning rate and weight decay.}
    \label{fig:place}
\end{figure}

We begin by conducting ablation studies to adjust the hyperparameters.
The results are shown in Fig.~\ref{fig:place}. By continuously adjusting the learning rate and comparing the scores of each task, we observe that only when using the learning rate set in Sec.~\ref{sec:experiment} could the indicators reach their optimum. 

\subsection{Impact of Data and Model Size}
We conduct additional experiments varying the data size and model size in Tab. \ref{tab:size}.
We use four datasets, namely RAVEN, CVR, SVRT and BongardHOI.
As the amount of data and model parameters increases, the performance of the reasoner improves.
This aligns with our \textit{approximation principle}, indicating that we can ultimately approximate a better reasoner with larger data volumes and model parameters.

\subsection{Comparison of Reasoning Performances of LLM-Based Models}
\begin{figure}[!t]
    \centering
    \includegraphics[width=0.58\textwidth]{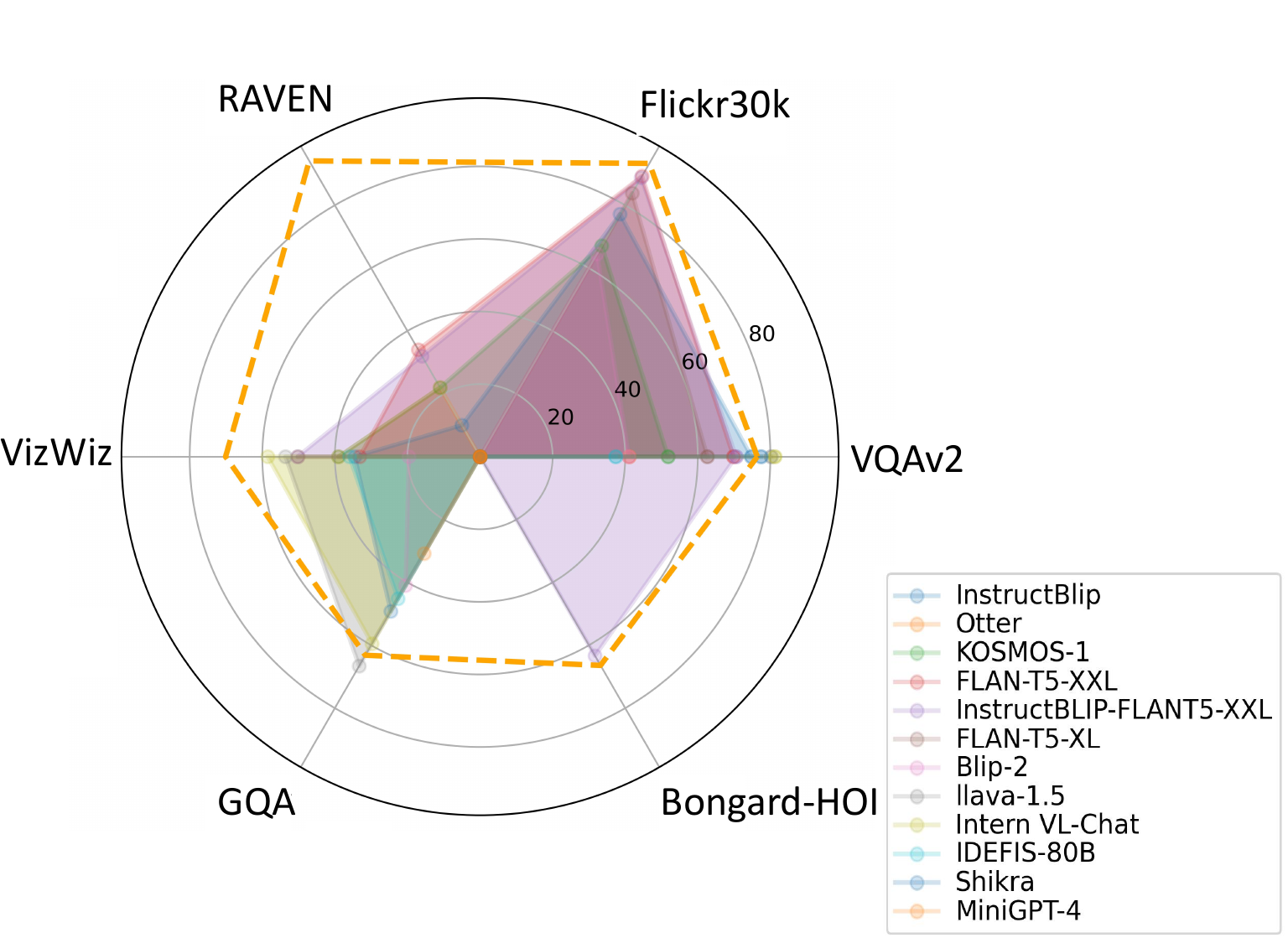}
    \caption{Performance of VLMs on visual reasoning. The orange dashed line indicates the performance of small ad-hoc models (non-VLMs)\cite{spratley2020closer,jiang2020defense,raghuraman2023crossimage,nguyen2023hada,wang2022git,kervadec2021supervising}. Some data points are missing due to the lack of publicly available data for testing the model on the related datasets. The ad-hoc models still outperform existing VLMs on visual reasoning tasks.}
    \label{fig:radar}
\end{figure}

We compare the ad-hoc models with LLM-based models~\cite{zhu2023minigpt, zhao2023mmicl} in Fig.~\ref{fig:radar}.
For RAVEN, LLMs all perform relatively poorly.
For VQA, our LLM-based model reaches a comparable score with others.
For Bongard-HOI, despite the efficiency of InstructBLIP-FLANT5-XXL (12.1B)~\cite{zhao2023mmicl}, we only test our LLM-based model on MiniGPT-4 with a small parameter size of 7B and without complex in-context-learning tricks.
It can be observed that most LLM-based models still struggle on visual reasoning problems.

\begin{table}[!t]
    \centering
    \caption{Impact of data size and model Size on model performance.}
    \label{tab:size}
    \begin{minipage}[t]{0.495\linewidth}
        \centering
        \resizebox{\linewidth}{!}{       
        \begin{tabular}{c|cccc}
        \toprule
            Data Size & \textbf{RAVEN} & \textbf{CVR} & \textbf{SVRT} & \textbf{BongardHOI} \\
        \midrule
            2000 & 18.09 & 45.96 & 50.38 & 50.14 \\
            6000 & 29.12 & 52.75 & 50.17 & 50.87 \\
            10000 & 36.28 & 76.85 & 50.63 & 52.15\\
            20000 & \textbf{38.02} & \textbf{77.84} & \textbf{51.43} & \textbf{52.22}\\
         \bottomrule
        \end{tabular}
        }
    \end{minipage}
    \begin{minipage}[t]{0.495\linewidth} 
        \centering
        \resizebox{\linewidth}{!}{
        
        \begin{tabular}{c|cccc}
        \toprule
            Model Size & \textbf{RAVEN} & \textbf{CVR} & \textbf{SVRT} & \textbf{BongardHOI} \\
        \midrule
            8.1k & 31.30 & 74.47 & 51.08 & 50.12\\
            16.2k & 31.29 & 75.95 & 51.39 & 50.89\\
            32.5k & 45.33 & 75.13 & 64.90 & \textbf{51.86} \\
            97.5k & \textbf{47.82} & \textbf{77.35} & \textbf{84.96} & 51.71\\
         \bottomrule
        \end{tabular}
        }
    \end{minipage} 
\end{table}

\begin{table*}[t]
    \centering
    \caption{Comparison of the results between an End-to-End model with our One-for-All considering different types of CVR questions. From the results, we can easily find the failure case of the One-for-All model, thus figuring out the principle for future network design.
    }
    \resizebox{0.85\textwidth}{!}{
    \begin{tabular}{c|ccccccccc}
        & Shape & Position & Size & Color & Rotation & Flip & Count & Inside & Contact  \\
        \midrule
        Ad-hoc End-to-End & \textbf{69.0} & \textbf{93.0} & 76.4 & 26.7 & \textbf{67.8} & \textbf{65.2} & \textbf{91.6} & \textbf{98.8} & \textbf{94.3}  \\
        One-for-All & 44.4 & 88.4 & \textbf{90.0} & \textbf{33.3} & 45.7 & 46.2 & 86.1 & 97.6 & 90.7 \\
    \end{tabular}
    }
    \label{tab:failure}
\end{table*}

\begin{figure*}[!t]
    \centering
    \includegraphics[width=\textwidth]{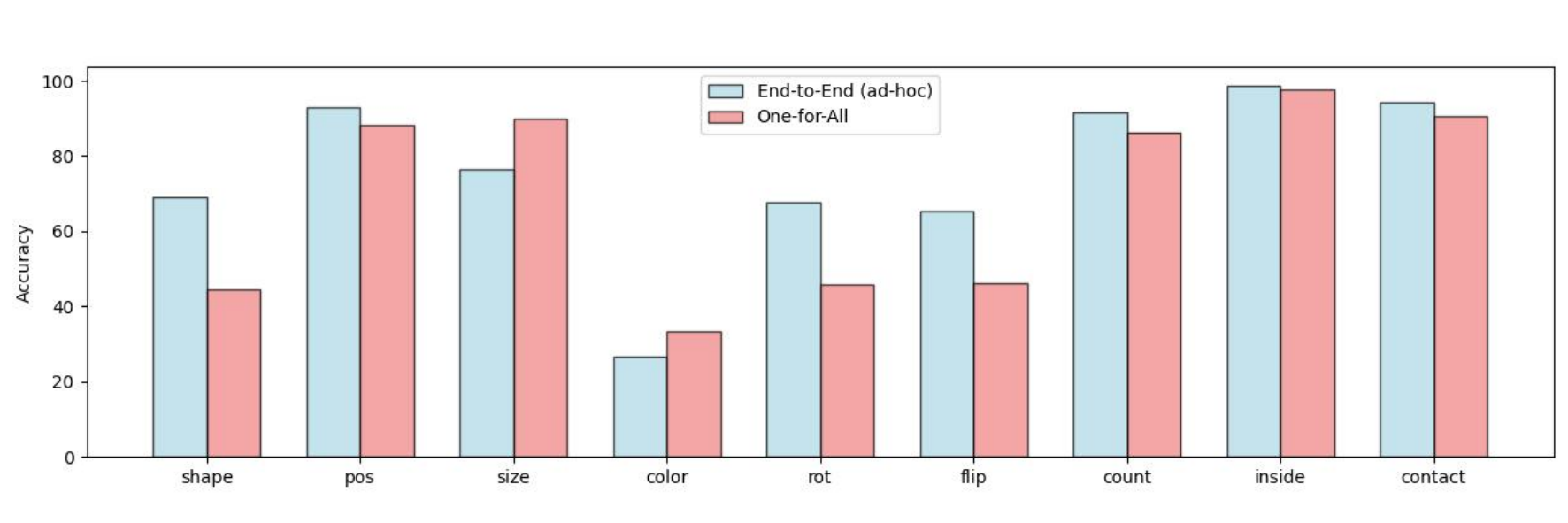}
    \caption{Comparison of the results between the ad-hoc End-to-End model with our One-for-All considering different types of CVR tasks.}
    \label{fig:CVRfail}
\end{figure*}

\subsection{Failure Case Analysis}

For CVR~\cite{zerroug2022benchmark} elementary problem, it contains nine tasks respectively focusing on shape, position, size, color, rotation, flip, count, inside, and contact.
We conduct failure case analysis on the CVR subset task with an ad-hoc End-to-End model and our One-for-All model.

As shown in Tab.~\ref{tab:failure} and Fig.~\ref{fig:CVRfail}, the two models show different performances on different tasks.
Overall, the One-for-All model performs slightly worse than the End-to-End trained model on the test dataset. 
This discrepancy can potentially be attributed to overfitting in the End-to-End model, which is specifically designed to tackle the CVR task.
For some tasks such as size and color, we notice that our One-for-All model outperforms the ad-hoc End-to-End model.
This can be attributed to the enhancement of performance in these tasks by the reasoner network through learning from other datasets.

%
%

\end{document}